\documentclass{article}
\usepackage{spconf,amsmath,graphicx,mathtools,nth,hhline,multirow}

\title{Hit Song Prediction for Pop Music by Siamese CNN with Ranking Loss}
\name{Lang-Chi Yu$^{*}$, Yi-Hsuan Yang$^{*}$, Yun-Ning Hung$^{*}$, Yi-An Chen$^{\dagger}$ \thanks{Copyright 2018 IEEE. Published in the IEEE 2018 International Conference on Acoustics, Speech, and Signal Processing (ICASSP 2018), scheduled for 15-20 April 2018 in Calgary, Alberta, Canada. Personal use of this material is permitted. However, permission to reprint/republish this material for advertising or promotional purposes or for creating new collective works for resale or redistribution to servers or lists, or to reuse any copyrighted component of this work in other works, must be obtained from the IEEE. Contact: Manager, Copyrights and Permissions / IEEE Service Center / 445 Hoes Lane / P.O. Box 1331 / Piscataway, NJ 08855-1331, USA. Telephone: + Intl. 908-562-3966.}}
\address{$^{*}$MAC Lab, Research Center for Information Technology Innovation, Academia Sinica, Taiwan \\
$^{\dagger}$Machine Learning Research Team, KKBOX Inc., Taiwan \\
{\small \tt \{yl871804,yang,biboamy\}@citi.sinica.edu.tw, annchen@kkbox.com}}

\begin{document}
%
\maketitle
\begin{abstract}
A model for hit song prediction can be used in the pop music industry to identify emerging trends and potential artists or songs before they are marketed to the public.  While most previous work formulates hit song prediction as a regression or classification problem, we present in this paper a convolutional neural network (CNN) model that treats it as a ranking problem. 
Specifically, we use a commercial dataset with daily play-counts to train a multi-objective Siamese CNN model with Euclidean loss and pairwise ranking loss to learn from audio the relative ranking relations among songs. 
Besides, we devise a number of pair sampling methods according to some empirical observation of the data. 
Our experiment shows that the proposed model with a sampling method called A/B sampling leads to much higher accuracy in hit song prediction than the baseline regression model. Moreover, we can further improve the accuracy by using a neural attention mechanism to extract the highlights of songs and by using a separate CNN model to offer high-level features of songs. 
\end{abstract}
\section{Introduction}
\label{sec:intro}

Hit song prediction, as defined by Pachet \textit{et al.}~\cite{pachet2012hit}, aims at ``predicting the success of songs before they are released to the market.'' One of its possible applications is to help music streaming companies to identify new songs, artists, and emerging trends worth investing and promoting
\cite{celma2009music}. 
Another application is to use a hit song prediction model along
with a human composer or an automatic composition model to generate new hits or improve existing ones \cite{herremans2014dance}. 

This paper is an extension of a previous work~\cite{yang2017revisiting} that used deep learning for hit song prediction from audio, which had been rarely attempted in the literature. Many previous works viewed hit song prediction as a regression (or rating) or classification problem with various approaches, including SVM classifiers based on latent topic features from audio and lyrics~\cite{dhanaraj2005automatic} or on human-annotated tags~\cite{pachet2008hit}, Bayesian network based on lyric features only~\cite{singhi2014hit}, and time weighted
linear regression~\cite{fan2013study}. Different input features of songs other than audio and lyrics were also used, e.g. play-counts extracted from \#nowplaying Twitter tweets~\cite{kim2014nowplaying,zangerle2016can}. Yang \textit{et al.}~\cite{yang2017revisiting} first applied deep learning approach to hit song prediction. In \cite{yang2017revisiting}, 
each song was associated with a \emph{hit score} indicating its popularity in the market. Various regression models, including linear regression, plain CNN, and Inception CNN 
\cite{szegedy2015going} were used to predict the hit scores. Experiment showed that deeper model performs much better than shallower models. 

\begin{figure}[t!]
\centering
\includegraphics[width=.5\linewidth]{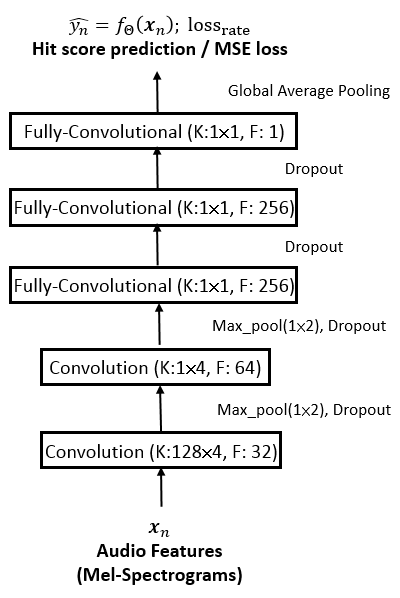}
\caption{A simple CNN model for hit song prediction (rating);  $K$ denotes kernel size and $F$ the number of filters.}
\label{fig:simp_cnn_arch}
\end{figure}

\begin{figure*}[t!]
  \centering
  \includegraphics[width=.28\textwidth]{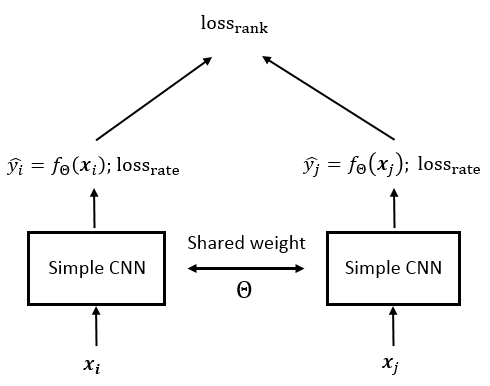}
  \includegraphics[width=.28\textwidth]{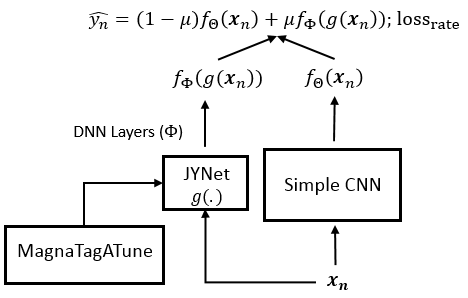}
  \includegraphics[width=.4\textwidth]{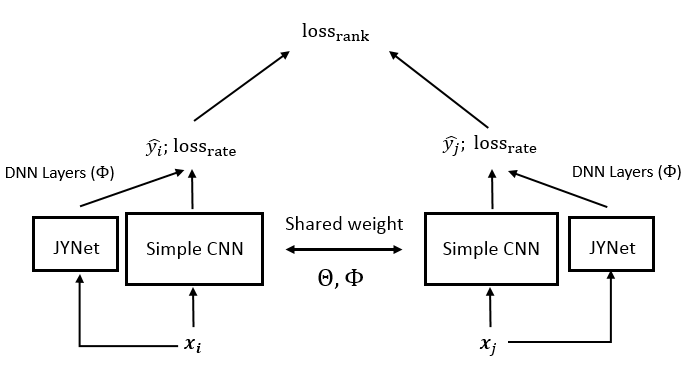}
\\
  (a) Siamese CNN ~~~~~~~~~~~~~~ 
  (b) Simple CNN with tag features~~~~~~~~~~~~~~
  (c) Siamese CNN with tag features
\caption{Proposed models}
\label{fig:proposed_models}
\end{figure*}

We propose to extend~\cite{yang2017revisiting} in several aspects. Foremost, since the main goal of hit song prediction is to tell hits from non-hits, learning relative popularity or ranking among songs might be sufficient and more plausible, since the prediction of the actual play-count may be biased by many other factors. Following this light, we propose to use a Siamese architecture~\cite{chopra2005learning} that optimizes the model parameters by jointing considering rating- and ranking-based loss functions.
Next, as the input to a Siamese network is pairs of songs, we propose three different methods to form the song pairs to train the model with different combinations of data. One of the method turns out to be effective in mitigating the imbalance between unpopular and popular songs in the training set.
Third, we study whether additional tag-based input features  help improve the accuracy of the Siamese model. 
Fourth, instead of using cumulative song play-counts over a long time span as did in~\cite{yang2017revisiting}, we propose (with rationals) to use the daily play-count of a song on a certain date after its first release to the market to define its hit score. Finally, a music thumbnailing model~\cite{huang2017music} is used to help our model extract more representative audio representation of songs.


In what follows, we  firstly present the proposed models in Section~\ref{sec:models}, the dataset in Section~\ref{sec:dataset}, and finally the experimental setup and results in Section~\ref{sec:exp_results}. We conclude  in Section~\ref{sec:conclusion}.

\section{Models}
\label{sec:models}

The proposed model is a multi-objective Siamese CNN which takes song pairs as input and jointly optimizes both the mean squared error (MSE) in predicting (i.e. \emph{rating}) the hit song scores for both songs and the pairwise \emph{ranking} loss in deciding which one of the two is likely to have a higher score. Below, we firstly present the baseline rating-only CNN model, and then several extensions that we add to it.

\subsection{Simple CNN}
\label{ssec:simple_cnn}

Given the groundtruth hit score $y_n$ (see Section \ref{sec:dataset} for how we define it) and the low-level feature representation $\textbf{x}_n$ (see Section \ref{ssec:music_thumbnailing}) for each song $n$ in the training set, we build a CNN model $f(\cdot)$ for hit song rating by finding the optimal parameter set $\Theta^{*}$ that has the minimal empirical MSE, i.e.:
\begin{equation} \label{eq:cnn_obj}
\Theta^{*} = \arg\min_{\Theta} \text{loss}_{\text{rate}} = 
\arg\min_{\Theta} \frac{1}{N}\sum_{n} \| y_n - f_{\Theta}(\textbf{x}_n) \|_{2}^{2} \,,
\end{equation}
where $N$ denotes the number of songs in the training set. 
With $f_{\Theta^{*}}(\cdot)$, we can still rank the songs in the test set in terms of the predicted hit scores, despite that the model itself is optimized for rating not for ranking.

In our implementation, we used the network architecture that has been shown effective in hit score rating in \cite{yang2017revisiting}, with two convolutional layers and three fully-convolutional layers. The details are depicted in Fig.~\ref{fig:simp_cnn_arch}. 

\subsection{Multi-objective Siamese CNN}
\label{ssec:siamese_cnn}


We can train a Siamese CNN instead by directly optimizing a ranking-based loss. As shown in Fig.~\ref{fig:proposed_models} (a), a Siamese CNN is composed of two identical simple CNN models, which share the same parameter set $\Theta$. 
Given input pairs $(\textbf{x}_i, \textbf{x}_j)$ from the training set, we optimize $\Theta$ by minimizing the following pairwise ranking loss proposed in \cite{kong2016photo}:
\begin{equation} \label{eq:rank_loss}
\text{loss}_\text{rank} = \frac{1}{P}\sum_{i,j}\max(0, m-\delta(y_i,y_j)(f_{\Theta}(\textbf{x}_i)-f_{\Theta}(\textbf{x}_j)))\,,
\end{equation}
where $P$ denotes the number of song pairs we use, $m>0$ is called the ``margin'' and is a hyper-parameter to be tuned by using the validation set, and $\delta(y_i,y_j)$ returns $1$ if $y_i\geq~y_j$ and $-1$ otherwise. Therefore, if $y_i\geq~y_j$, Eq. \ref{eq:rank_loss} prefers that $f_{\Theta}(\textbf{x}_i)-f_{\Theta}(\textbf{x}_j)\geq m$. It encourages the model to rank the songs correctly (with certain confidence that is related to $m$), without considering the actual difference between $y_i$ and $y_j$.

We can combine the two loss functions and have a multi-objective Siamese CNN $f_{\Theta}(\cdot)$ that minimizes
\begin{equation} \label{eq:mse_rank_loss}
\text{loss}_\text{multi} = (1-w)~\text{loss}_\text{rate} + w ~\text{loss}_\text{rank}\,,
\end{equation}
where $w\in[0,1]$ is another hyper-parameter to be decided from the validation data.

\subsection{Selection of Data Pairs}
\label{ssec:siamese_data}

For the specific task of hit song prediction, we find it important to investigate how to form the data pairs for the Siamese CNN and propose the following three methods. 
First, we can just na\"ively sample random song pairs (without replacement) from the whole training set. This is called \emph{na\"ive sampling}. 

Second, as we will elaborate in Section \ref{sec:dataset}, there is usually a so-called ``long-tail'' in music listening data, meaning that people usually listen to only a small subset of songs \cite{celma2009music}. With na\"ive sampling, we might have a large number of song pairs that are composed of two unpopular songs. To counter this data imbalance issue, we propose the \emph{A/B sampling} method that firstly divides the training set into two groups A and B, depending on whether the hit score of a song is greater than the average hit score of the whole training set, and then requires every sampled song pair to have at least one song from group A (the popular one). In this way, we avoid comparing the hit scores of songs that are in the long tail.

With information regarding the artist(s) who performed a song, we also employ an \emph{artist sampling} method that requires the two songs in a pair to be from the same artist(s).  In this way, our model is forced to figure out why some songs from an artist is popular but some others are not.

The A/B sampling and artist sampling methods may have respectively a global (inter-artist) and local (intra-artist) sense of popularity. Therefore, we also try to combine them by taking the average of the hit scores they estimate.


\subsection{Training with High Level Features}
\label{ssec:music_feat}

Generic low-level features of songs, like the mel-spetrogram used in \cite{yang2017revisiting} and also throughout this work, may suffer from the ``semantic gap'' \cite{dorai2003bridging} and cannot lead to an accurate prediction model for a high-level concept such as hotness. To address this issue, we make use of an external dataset with semantically rich tags (labels) such as genres and instruments, to train a multi-label music tagging CNN model. Then we feed the songs of our dataset to this tagging model and take the output as an additional input feature representation to train a hit score rating or ranking CNN model. 

In our implementation, we use JYnet~\cite{liu2016event}, a pre-trained music tagging CNN model that learns to predict the activation scores (from $0$ to $1$) of 50 music tags 
based on the MagnaTagATune dataset \cite{law2009evaluation}, which is composed of mainly Western Pop music. 
As shown in Figs.~\ref{fig:proposed_models} (b) and (c), the predicted activation scores are treated as 50-dimensional music features and passed through a 3-layer fully-connected DNN $f_{\Phi}(.)$ to learn to predict the hit scores. 
We used hidden layer sizes 100, 100, 30 respectively for the three layers and jointly optimized the parameter set $\Phi$ for this tag-based DNN model and the parameter set $\Theta$ for the aforementioned audio-based CNN model by taking the weighted combination of their outputs. 
The loss function becomes, for example for the rating-only model (i.e. the one shown in Fig. \ref{fig:proposed_models} (b)), 
\begin{equation} \label{eq:cnn_hl_feat_mse_loss}
\frac{1}{N}\sum_{n} \| y_n - (1-\mu) f_{\Theta}(\textbf{x}_n) - \mu f_{\Phi}(g(\textbf{x}_n)) \|_{2}^{2}\,,
\end{equation}
where $\mu \in [0,1]$ is a hyper-parameter to be decided from the validation set and $g(\cdot)$ is the pre-trained music tagging model.

\subsection{Music Thumbnailing and Audio Features}
\label{ssec:music_thumbnailing}

The generic low-level audio features extracted from songs (i.e. $\mathbf{x}_n$) are 128-bin log-scaled mel-spectrograms~\cite{dieleman2013multiscale}, computed by using the librosa library~\cite{librosa}, with sampling rate 22,050 Hz and half-overlapping hamming window of size 4,096 samples for short-time Fourier transform. To control the temporal length of the model input, we compute the mel-spectrogram from a 30-second segment per song. 
There are two ways to get the segments. A na\"ive way is to take directly the middle 30-second segment, as did in ~\cite{yang2017revisiting}. We call this the \emph{Mid-30} method. Alternatively, \emph{HL-30} uses a state-of-the-art neural attention model for music thumbnailing~\cite{huang2017music} (details omitted due to space restriction) to extract a 30-second highlight for each song. We will empirically compare the performance of the two methods in the experiment.

\begin{figure}[t!]
  \hspace{0.3cm}
\begin{minipage}[b]{.45\linewidth}
  \centering
  \includegraphics[width=4.5cm]{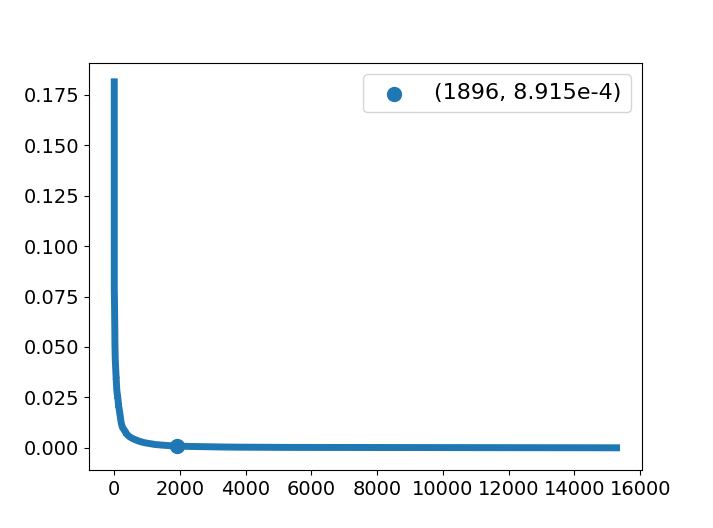}
  \centerline{(a) Training+validation+test set}\medskip
\end{minipage}
\hspace{0.3cm}
\begin{minipage}[b]{.45\linewidth}
  \centering
  \centerline{\includegraphics[width=4.5cm]{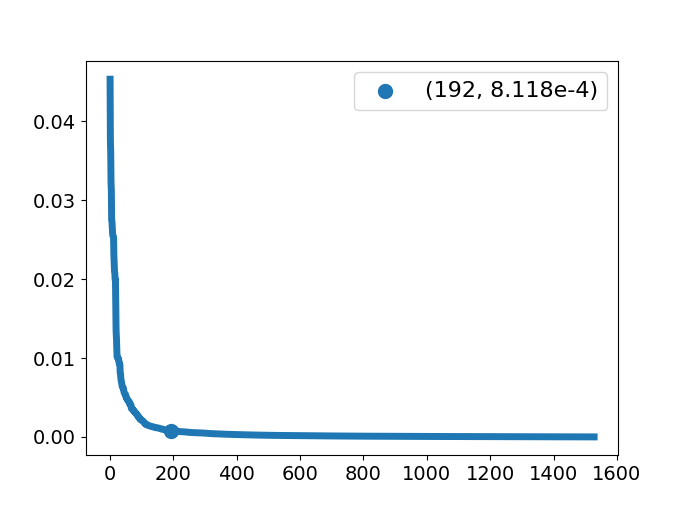}}
  \centerline{(b) Test set}\medskip
\end{minipage}
\caption{Distributions and average value occurrences of hit scores of KKBOX dataset}
\label{fig:dist_hit_scores}
\end{figure}


\section{Dataset}
\label{sec:dataset}

Our work is based on daily play-counts of pop song from the listening logs of KKBOX Inc., a leading music streaming service provider in Taiwan and East Asia. The dataset we used for training and evaluation is collected from Jan. 2016 to Jun. 2017, a span of one year and a half, involving about 30K users and 2M songs. Due to a confidentiality agreement with KKBOX, the daily play-count for a song represents the play-count of that song from all users that day \emph{normalized} (i.e. divided) by the total number of play-count of all the songs from all users that day.  In other words, it is the \emph{daily market share} of a song was used in place of its actual daily play-count.

\begin{table*}[th] 
\caption{\label{tab:results_simple_n_siam}Result of different models for hit score prediction; we use bold font to highlight the top two results.}
        \vspace{2mm}
        \centering
          \begin{tabular}{| c |l | l | l | c  c  c | }
            \hline
&network type              & sampling method & feature & nDCG@10\% & Kendall@10\% & Spearman@10\% \\ \hline \hline
               (a) &\multicolumn{2}{|l|}{} &
                 audio (Mid-30) & 0.0725 & -0.0421 & -0.0696  \\ 
                (b) &\multicolumn{2}{|l|}{Simple CNN}  &
                  audio (Mid-30) + tag & 0.0872 & 0.0157 & 0.0225 \\ \cline{4-7}
                (c) &\multicolumn{2}{|l|}{(rating only)}  &
                  audio (HL-30) & 0.0999 & 0.0735 & 0.1121  \\ 
               (d) &\multicolumn{2}{|l|}{} &
                  audio (HL-30)  + tag & 0.1241 & 0.1080 & 0.1663 \\ \cline{2-7}
                (e) && \multirow{2}{*}{na\"ive}  &
                  audio (HL-30) & 0.1069 & 0.0828 & 0.1222  \\ 
                (f) &&  &
                  audio (HL-30)  + tag & 0.0994 & 0.1059 & 0.1578 \\ \cline{3-3}
                 (g) &&
                \multirow{2}{*}{artist}  &
                  audio (HL-30) & 0.1200 & 0.0575 & 0.0854 \\ 
                (h) &Siamese CNN  &  &
                  audio (HL-30)  + tag & 0.1285 & 0.1687 & 0.2465 \\ \cline{3-3}
                (i) &(rating + ranking) &
                \multirow{2}{*}{A/B}  &
                  audio (HL-30) & 0.1127 & 0.1852 & 0.2713  \\ 
                 (j) && &
                  audio (HL-30)  + tag & \textbf{0.1287} & \textbf{0.2415} & \textbf{0.3481} \\ \cline{3-3}
                  (k) &&
                \multirow{2}{*}{A/B + artist}  &
                  audio (HL-30) & 0.1127 & 0.1868 & 0.2753  \\ 
                 (l) && &
                  audio (HL-30)  + tag & \textbf{0.1287} & \textbf{0.2421} & \textbf{0.3484} \\ \hline %
          \end{tabular}
\end{table*}

In addition to using daily play-counts instead of the cumulative play-counts (e.g. the total number of play-counts for a song over a long period of time) as did in \cite{yang2017revisiting},  we define in this paper the \emph{hit score} of a song as its (daily)  \emph{play-count on the \nth{60} day after the release of that song}. 
There are two reasons for making this choice.
First, cumulative play-count may be biased since songs released earlier are likely to receive more play-counts than those that are released later. 
Second, some songs may enjoy a burst in popularity due to some external factors such as advertisement or some special promotion. 
We think  if a song is indeed a hit, it should remain popular even it has been released for a while.

\section{Experimental Setup and Result}
\label{sec:exp_results}

From the whole dataset, we sampled 15K songs with highest hit score (as defined above) and used 10-fold cross validation for data splitting, taking 12k, 1.5k, and 1.5k songs (i.e. 8:1:1) as the training, validation, and test set, respectively.  In other words, a song can appear only in the training, validation or test set. Such a data split is done 10 times so that each  fold is used as the test set once. 
To gain insight into the data and avoid biased sampling, we plot the corresponding sorted hit scores of the whole 15k songs ((a)) and test set only ((b)) in Fig.~\ref{fig:dist_hit_scores}, where the x-axes indicates songs with the n-th highest hit scores, y-axes stands for the normalized daily play-count, and big dots mark the ranks and values of average hit scores. From Fig.~\ref{fig:dist_hit_scores}, it can be seen that the two distributions ((a) and (b)) of sampled data are similar, and that the distributions are quite imbalanced with the play-count of about $87.5\%$ of songs are below the average. Though the test set may not be biased-sampled, data imbalanced problem is crucial and must be dealt with. The A/B sampling method presented in Section~\ref{ssec:siamese_data} may be a solution to this problem. 

For evaluation, the following evaluation metrics are considered: normalized discounted cumulative gain (nDCG@10\%), Kendall's $\tau$ (Kendall@10\%), and Spearman's $\rho$ (Spearman@10\%) for songs in the top 10\% of test set (i.e. top 150) with regard to true hit scores.
The nDCG takes both ranking positions and relevance scores (i.e. actual hit scores) into consideration, while Kendall's $\tau$ and Spearman's $\rho$ are only based on relative ranking positions. 
The result is averaged across the 10 iterations of data splitting.

The result is shown in Table~\ref{tab:results_simple_n_siam}. The upper half is for the simple, rating-only CNN models described in Section~\ref{ssec:simple_cnn}, where two methods for taking 30-second segments, Mid-30 and HL-30 presented in Section~\ref{ssec:music_thumbnailing}, were applied. Compared rows (c)(d) to rows (a)(b), we see that segmenting input with HL-30 yields much better result than with Mid-30, suggesting that the music thumbnailing model can generate highlights that carry more representative information of the songs. We find HL-30 also outperform Mid-30 for the Siamese CNN models and discuss only the result of HL-30 below.

The lower half of Table~\ref{tab:results_simple_n_siam} is for Siamese CNN models presented in Section~\ref{ssec:siamese_cnn}, using the three different pair sampling methods presented in Section~\ref{ssec:siamese_data}.
Results show that while artist sampling does not guarantee overall improvement 
(rows (g)(h)) as compared with na\"ive sampling, A/B sampling does perform pretty well as compared with the other two methods, especially in terms of Kendall's $\tau$ and Spearman's $\rho$. 

Table~\ref{tab:results_simple_n_siam} also shows that adding JYnet tags 
improves the result for almost all models. Siamese CNN model with A/B sampling and  tags (row (j)) produces the best result among all models. Fusing the result of A/B sampling and artist sampling (with weights also empirically determined from the validation set) improves the result slightly but not much.

\section{Conclusion}
\label{sec:conclusion}

In this paper, we have presented a Siamese CNN model based on state-of-the-art deep learning techniques from previous work on audio-based hit song prediction. While previous works usually formulate hit song prediction as a regression problem, our model learns both hit scores and relative ranking of songs jointly. Evaluation on daily play-counts of songs from the commercial data provided by KKBOX confirms that the Siamese structures are more effective than simple counterparts in discriminating hits from non-hits.
In addition, various pair sampling techniques can be used in Siamese structure to deal with data-specific problems; for example,  A/B sampling alleviates data imbalance problem in our experiments. Interpolation results of A/B sampling and artist sampling show possibility for future work to incorporate artist information into model design.




\bibliographystyle{IEEEbib}
\bibliography{refs}

\begin{thebibliography}{10}

\bibitem{pachet2012hit}
Fran{\c{c}}ois Pachet and CSL Sony,
\newblock ``Hit song science,''
\newblock {\em Music data mining}, pp. 305--26, 2012.

\bibitem{celma2009music}
{\`O}scar Celma et~al.,
\newblock {\em Music recommendation and discovery in the long tail},
\newblock Universitat Pompeu Fabra, 2009.

\bibitem{herremans2014dance}
Dorien Herremans, David Martens, and Kenneth S{\"o}rensen,
\newblock ``Dance hit song prediction,''
\newblock {\em J. New Music Research}, vol. 43, no. 3, pp. 291--302, 2014.

\bibitem{yang2017revisiting}
Li-Chia Yang, Szu-Yu Chou, Jen-Yu Liu, Yi-Hsuan Yang, and Yi-An Chen,
\newblock ``Revisiting the problem of audio-based hit song prediction using
  convolutional neural networks,''
\newblock in {\em Proc. Int. Conf. Acoustics, Speech, and Signal Processing},
  2017.

\bibitem{dhanaraj2005automatic}
Ruth Dhanaraj and Beth Logan,
\newblock ``Automatic prediction of hit songs,''
\newblock in {\em Proc. Int. Soc. Music Information Retrieval Conf.}, 2005, pp.
  488--491.

\bibitem{pachet2008hit}
Fran{\c{c}}ois Pachet and Pierre Roy,
\newblock ``Hit song science is not yet a science.,''
\newblock in {\em Proc. Int. Soc. Music Information Retrieval Conf.}, 2008, pp.
  355--360.

\bibitem{singhi2014hit}
Abhishek Singhi and Daniel~G Brown,
\newblock ``Hit song detection using lyric features alone,''
\newblock {\em Proc. Int. Soc. Music Information Retrieval Conf.}, 2014.

\bibitem{fan2013study}
Jianyu Fan and Michael~A Casey,
\newblock ``Study of {Chinese} and {UK} hit songs prediction.,''
\newblock {\em Proc. Computer Music Multidisciplinary Research}, 2013.

\bibitem{kim2014nowplaying}
Yekyung Kim, Bongwon Suh, and Kyogu Lee,
\newblock ``\# nowplaying the future billboard: mining music listening
  behaviors of twitter users for hit song prediction,''
\newblock in {\em Proc. Int. Workshop on Social Media Retrieval and Analysis}.
  ACM, 2014, pp. 51--56.

\bibitem{zangerle2016can}
Eva Zangerle, Martin Pichl, Benedikt Hupfauf, and G{\"u}nther Specht,
\newblock ``Can microblogs predict music charts? an analysis of the
  relationship between\# nowplaying tweets and music charts.,''
\newblock in {\em Proc. Int. Soc. Music Information Retrieval Conf.}, 2016.

\bibitem{szegedy2015going}
Christian Szegedy, Wei Liu, Yangqing Jia, Pierre Sermanet, Scott Reed, Dragomir
  Anguelov, Dumitru Erhan, Vincent Vanhoucke, and Andrew Rabinovich,
\newblock ``Going deeper with convolutions,''
\newblock in {\em Proc. IEEE Conf. Computer Vision and Pattern Recognition},
  2015, pp. 1--9.

\bibitem{chopra2005learning}
Sumit Chopra, Raia Hadsell, and Yann LeCun,
\newblock ``Learning a similarity metric discriminatively, with application to
  face verification,''
\newblock in {\em Prof. IEEE Computer Society Conf. Computer Vision and Pattern
  Recognition}, 2005, vol.~1, pp. 539--546.

\bibitem{huang2017music}
Yu-Siang Huang, Szu-Yu Chou, and Yi-Hsuan Yang,
\newblock ``Music thumbnailing via neural attention modeling of music
  emotion,''
\newblock in {\em Proc. Asia Pacific Signal and Information Processing
  Association Annual Summit and Conference}, 2017.

\bibitem{kong2016photo}
Shu Kong, Xiaohui Shen, Zhe Lin, Radomir Mech, and Charless Fowlkes,
\newblock ``Photo aesthetics ranking network with attributes and content
  adaptation,''
\newblock in {\em Proc. European Conf. Computer Vision}. Springer, 2016, pp.
  662--679.

\bibitem{dorai2003bridging}
Chitra Dorai and Svetha Venkatesh,
\newblock ``Bridging the semantic gap with computational media aesthetics,''
\newblock {\em IEEE MultiMedia}, vol. 10, no. 2, pp. 15--17, 2003.

\bibitem{liu2016event}
Jen-Yu Liu and Yi-Hsuan Yang,
\newblock ``Event localization in music auto-tagging,''
\newblock in {\em Proc. ACM Int. Conf. Multimedia}, 2016, pp. 1048--1057.

\bibitem{law2009evaluation}
Edith Law, Kris West, Michael~I Mandel, Mert Bay, and J~Stephen Downie,
\newblock ``Evaluation of algorithms using games: The case of music tagging.,''
\newblock in {\em Proc. Int. Soc. Music Information Retrieval Conf.}, 2009, pp.
  387--392.

\bibitem{dieleman2013multiscale}
Sander Dieleman and Benjamin Schrauwen,
\newblock ``Multiscale approaches to music audio feature learning,''
\newblock in {\em Proc. Int. Soc. Music Information Retrieval Conf.}, 2013, pp.
  116--121.

\bibitem{librosa}
Brian McFee, Matt McVicar, Oriol Nieto, Stefan Balke, Carl Thome, Dawen Liang,
  Eric Battenberg, Josh Moore, Rachel Bittner, Ryuichi Yamamoto, and et~al.,
\newblock ``librosa 0.5.0,''
\newblock Feb 2017.

\end{thebibliography}

\end{document}